\documentclass{article}
\usepackage{ijcai19}

\usepackage{times}
\usepackage{soul}
\usepackage{url}
\usepackage[utf8]{inputenc}
\usepackage[small]{caption}
\usepackage{graphicx}
\usepackage{amsmath}
\usepackage{booktabs}
\usepackage{algorithm}
\usepackage{algorithmic}
\usepackage{amssymb}

\def\ie{{\em i.e.},\ }

\title{Simultaneous Prediction Intervals for Patient-Specific Survival Curves}

\author{
Samuel Sokota\footnote{These authors contributed equally to this work.}\and
Ryan D'Orazio$^{\ast}$\and
Khurram Javed\and
Humza Haider\and
Russell Greiner\\
\affiliations
Department of Computing Science,
University of Alberta
\emails
\{sokota, rdorazio, kjaved, hshaider, rgreiner\}@ualberta.ca
}

\begin{document}

\maketitle

\begin{abstract}
Accurate models of patient survival probabilities provide important
information to clinicians prescribing care for life-threatening and 
terminal 
ailments.
A recently developed class of models -- known as individual 
survival
distributions (ISDs) -- produces
patient-specific survival functions that offer greater descriptive 
power of patient outcomes than was previously possible.
Unfortunately, at the time of writing, ISD models almost
universally lack uncertainty quantification.
In this paper we demonstrate that an existing method for 
estimating simultaneous
prediction intervals from samples can easily 
be adapted for patient-specific
survival curve analysis and yields accurate results.
Furthermore, we introduce both
a modification to the existing method and a novel method 
for estimating simultaneous prediction intervals
and show that they offer competitive performance.
It is worth emphasizing that 
these methods are not limited to survival analysis and can be
applied in any context in which sampling the distribution
of interest is tractable. Code is 
available at \url{https://github.com/ssokota/spie}.
\end{abstract}

\section{Introduction} 
\label{sec:Intro}
Understanding a patient's probability of survival
is critical for patient care -- 
predicting survival outcomes both facilitates better treatment
decisions and more informed time management.
While learning survival times resembles standard regression,
it is more challenging in that   
 many training instances are censored
 -- only a lower bound of the survival time is known.
Survival analysis, a field of statistics concerned with analyzing the
time until an event of interest occurs, offers a natural language for
studying patient survival times.
Unfortunately, most standard survival analysis tools 
cannot answer comprehensive questions about individual 
patients' survival probabilities.
Risk scores, such as those given by the Cox proportional
hazards model~\cite{cox}, produce patient-specific  hazard scores, 
which predict the order of patient deaths rather than survival 
probabilities.
Single-time probabilistic models, such as the Gail 
model~\cite{gail}, produce a probability of survival for each patient, but only
for one point in time.
Population-based survival curves, 
such as Kaplan-Meier~\cite{Kaplan-Meier}, 
offer probability values for each point in time but for populations
rather than individual patients.

\begin{figure}
\centering
  \includegraphics[width=\columnwidth]{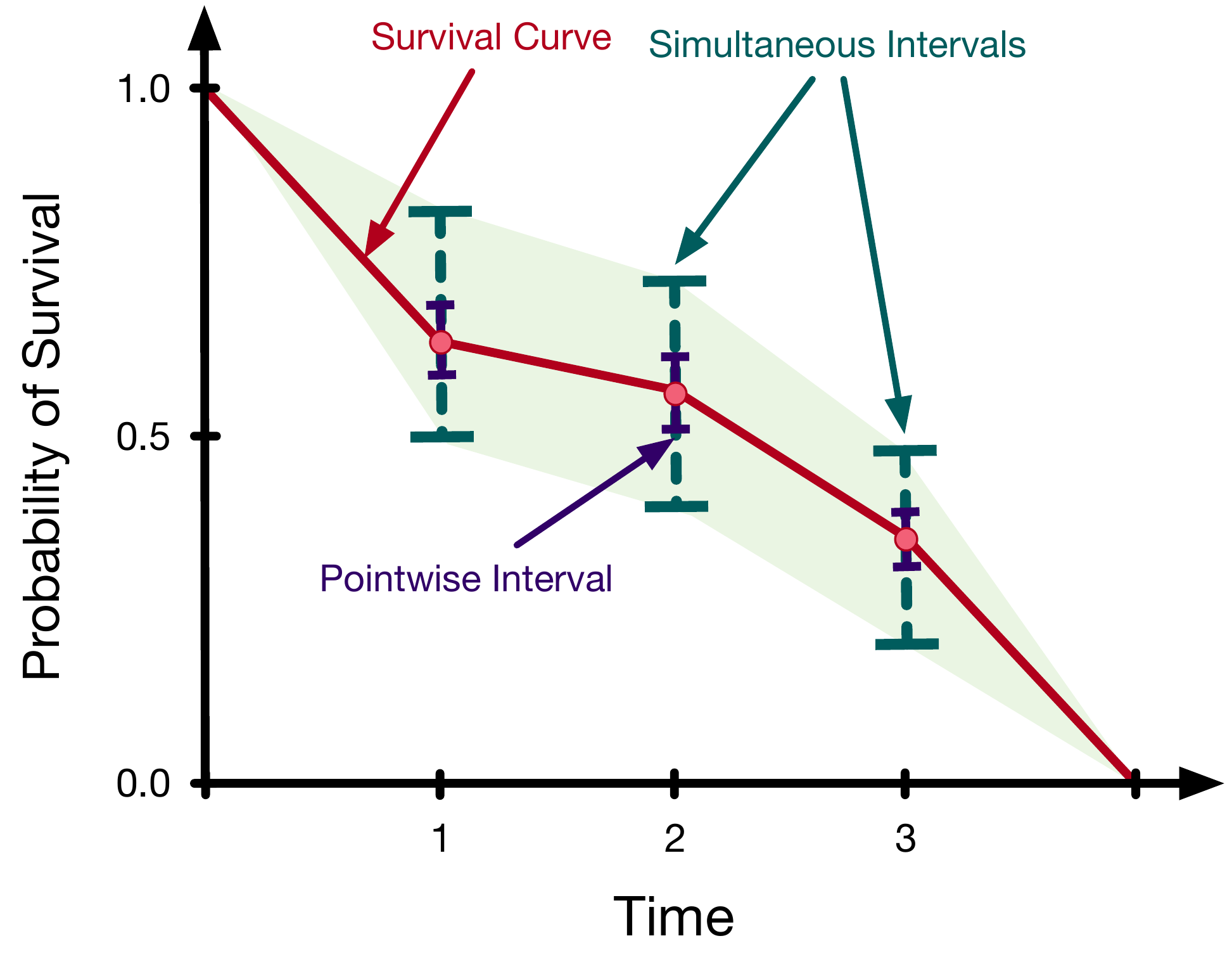}
  \caption{A survival curve (red line) gives survival probability as a function
  of time. The uncertainty of survival curves can be
  quantified by pointwise intervals, each of which covers its 
  corresponding value with the prescribed coverage probability,
  or by simultaneous intervals, 
  all of which simultaneously cover their corresponding
  value with the prescribed coverage probability. In the case that 
  simultaneous intervals exist for every time point we call the collection of intervals a region (light green area).}
  \label{fig:ISD}
\end{figure}

\begin{figure*}
\centering
  \includegraphics[width=\linewidth]{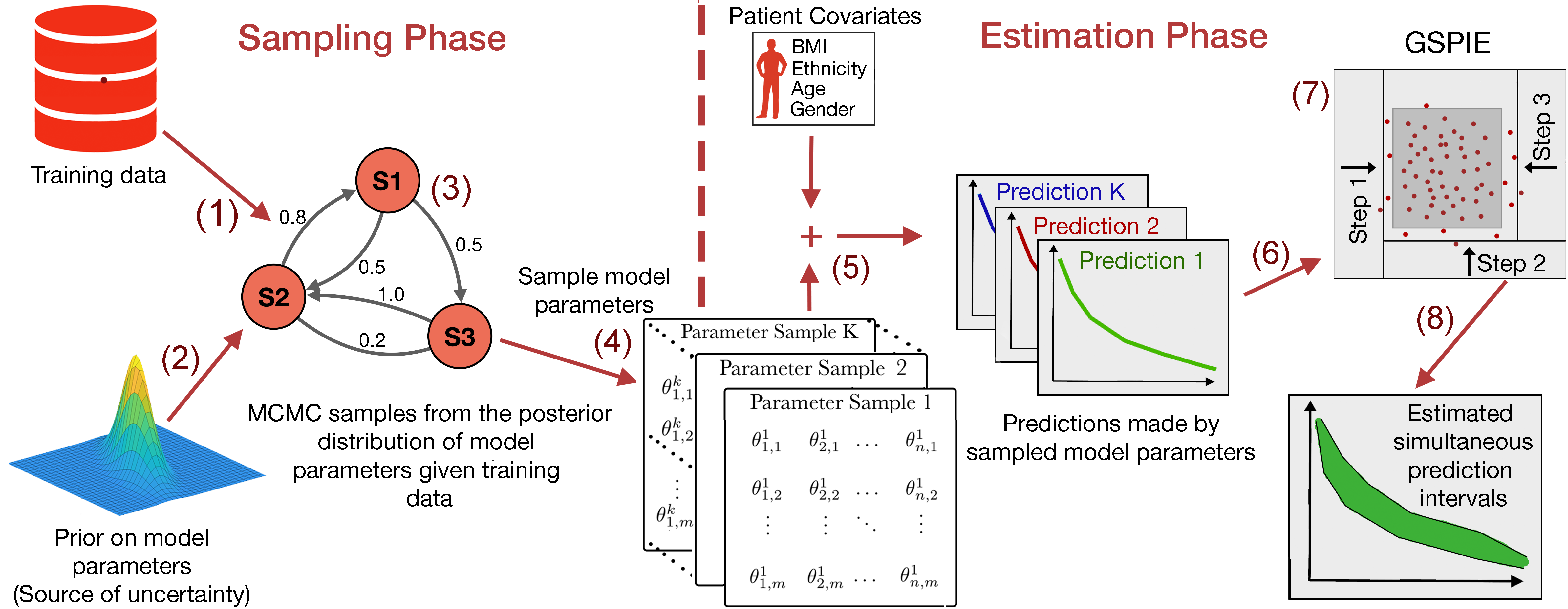}
  \caption{An example of the pipeline examined in this work. In the sampling phase (left)
  we acquire sample model instances that approximately reflect the uncertainty
  present in the system. While the figure shows this step for parametric posterior sampling, 
  this can also be done with non-parametric posterior sampling or bootstrapping.
  In the estimation phase (right) we estimate simultaneous 
  prediction intervals for new patients.
  Specifically, we map a given new patient to a survival curve with
  each sample model instance. Then using a SPIE,
  such as GSPIE, Olshen's method, or two-sided Olshen's method, we estimate simultaneous prediction intervals.}
\label{fig:inference}
\end{figure*}

These limitations have motivated a class of models that
learn an \textit{individual survival distribution} (ISD)\ -- 
a function
that gives the probability $P(T > t \mid \mathbf{x})$ 
that a patient with feature vector $\mathbf{x}$
will survive to at least time $t$~\cite{survey}. 
In other words, an ISD model 
produces a survival curve (see Figure \ref{fig:ISD}; red line)
specific to each patient.
This class of models
includes classical systems such as the the Cox model with the 
Kalbfleisch-Prentice extension (Cox-KP)~\cite{cox-cp-aft}, and
the Accelerated Failure Time model (AFT)~\cite{cox-cp-aft}, but also more 
recent 
models such as Random Survival Forests ~\cite{random-forests}, Multi-Task 
Logistic Regression (MTLR)~\cite{mtlr}, and a host of deep learning 
models~\cite{deepbengio,deepkatz,deeprag}.

Unfortunately, while the greater complexity of ISD models 
allows for greater expressive power,
it makes analytical uncertainty quantification difficult or impossible.
This is problematic because, to make an intelligent 
decision based on an agglomeration
of sources, it is essential that a clinician be able to understand
the trustworthiness of each piece of information. 
If clinicians cannot judge the reliability of patient-specific
survival curves, the applicability of ISD models in clinical settings
is limited.

In this work we seek to remedy this situation by showcasing 
an easily implementable pipeline for
estimating the uncertainty of patient-specific survival curves.
This pipeline has two phases. First, in the sampling 
phase, we acquire model samples using either bootstrapping or
posterior sampling. Second, in the estimation phase, we use
the model samples to estimate uncertainty for specific patients. 
Whereas analytical methods for estimating uncertainty are typically
model specific, this pipeline
is largely model agnostic -- it is applicable to
any model for which it is feasible to acquire bootstrap 
or posterior samples.
Many modern ISD models meet this condition and stand to benefit from
the advent of flexible, efficient, and accurate uncertainty quantification.

\section{Related Work} 
\label{sec:Related}

There is a large body of literature regarding the construction
of confidence intervals for population-based models.
The Kaplan-Meier (KM) estimator is an example of a 
model possessing both pointwise confidence 
intervals (see Figure \ref{fig:ISD}; blue lines), which can be derived 
with Greenwood's 
approximation coupled with 
the normal approximation~\cite{Greenwood}, and confidence regions
(see Figure \ref{fig:ISD}; light green area),
which can be derived with an adaptation of Greenwood's 
formula~\cite{nairconfidence} or using Brownian bridge limits
~\cite{hallconfidence}.
Approximate confidence regions also exist for the Cox 
model~\cite{linconfidence}, 
and a large class of semi-parametric 
transform models~\cite{chengpredicting} 
via approximations with a Gaussian process. 
The transformation models
allow for a non-parametric baseline hazard and relax the proportional 
hazard assumption, making them more flexible than the Cox model.
However, these transformation models assume that a predetermined function 
of the survival curve varies linearly with patient features.
Additionally, these approximations tend to be poor
at extreme time points, yielding useful confidence regions
only between the smallest and largest observed event times.

Individual survival distribution models are 
more expressive than population-based models but 
in general lack uncertainty quantification. 
In this work we address this shortcoming by 
proposing a sampled-based pipeline for
finding simultaneous prediction intervals
over a discrete sequence of time points
for patient-specific survival curves produced by ISD models.
The discretization of time is 
less of an issue than it might seem, as
many of the top performing ISD models already discretize time to
facilitate learning~\cite{mtlr,giunchiglia2018rnn,deepbengio}.
Moreover, for ISD models that produce a continuous prediction,
simultaneous prediction intervals 
can be estimated over a very large number of time points.

Closest to the present work, \citeauthor{olshen}~introduced
methodology to estimate simultaneous prediction intervals for 
gaits of normal children.
The methodology, henceforth Olshen's method, is an instance of
what we refer to as a simultaneous prediction intervals estimator (SPIE).
In this work we show that Olshen's method
can easily be applied to patient-specific survival analysis and offers
strong performance.
We also introduce both a modification to Olshen's
method and a novel SPIE which offer competitive performance.

\section{Background} 
\label{sec:Background}

An ISD is a function $f(\mathbf{x}, t) := P(T > t \mid \mathbf{x})$
inducing a survival curve for each patient with feature vector
$\mathbf{x} \in \mathbb{R}^d$.
In practice, many algorithms model an ISD 
over a discretized set of $n$ time points $t_1 < t_2 < \dots < t_n$
-- \ie each survival curve is
 a member of the $n$-discretized survival space
\[
\mathbb{S}^n := \{b \in [0, 1]^n : b_i \ge b_{i+1}\}.
\]
A learned ISD model $\mathbf{m}$ of $\mathbb{S}^n$ yields survival curves
given by
\[s^n \colon \mathbf{X} \times \mathbf{M} \to \mathbb{S}^n,\]
where
\[s^n(\mathbf{x}, \mathbf{m})_i \  := P_{\mathbf{m}}(T > t_i \mid \mathbf{x}).\]
For brevity, we will omit the $n$ superscript.

From a Bayesian perspective, $\mathbf{m}$
has prior distribution $p(\mathbf{m})$.
Because a patient's learned survival vector 
$s(\mathbf{x}, \mathbf{m})$ is a function
of $\mathbf{m}$, it is a random vector. 
This uncertainty is reflected by the 
predictive posterior distribution 
$p(s(\mathbf{x}, \mathbf{m}) \mid D)$
for observed training data $D$. 
Usually, the posterior distribution over models $p(\mathbf{m} \mid D)$ 
does not exist in closed form.
However, if the posterior can be computed up to a normalization constant
-- as is often the case --
it can be sampled using Markov chain Monte Carlo (MCMC) methods,
as is shown in the sampling phase of Figure \ref{fig:inference}.
Furthermore, if the gradient of the log posterior is available, 
modern MCMC methods~\cite{nuts} can efficiently sample high-dimensional
parameter spaces with minimal tuning~\cite{pymc}. 
For many models, this gradient exists and can be
computed with little effort using open source packages and automatic
differentiation~\cite{autodiff}.

From a frequentist perspective, uncertainty arises from the 
distribution over possible training datasets,
each of which may induce a different instance of the model. 
Since the empirical distribution of data 
approximates the underlying
distribution from which it was drawn,
sampling from the empirical distribution -- 
bootstrapping -- approximates sampling possible
training datasets from the underlying distribution. 
For models with efficient fitting procedures, an instance can be
fit to each bootstrapped dataset, 
yielding a set of
models analogous to Bayesian posterior samples collected from
MCMC methods.

Ultimately, we are interested in using these sampled models
to estimate simultaneous prediction intervals for the survival curves
of specific patients.
In the context of this work, a prediction interval is
an interval within which, with a certain probability, 
the patient's likelihood of survival falls at a 
certain time point given the observed information.
Prediction intervals arise in both Bayesian and
frequentist frameworks.
\begin{figure}
\centering
  \includegraphics[width=\columnwidth]{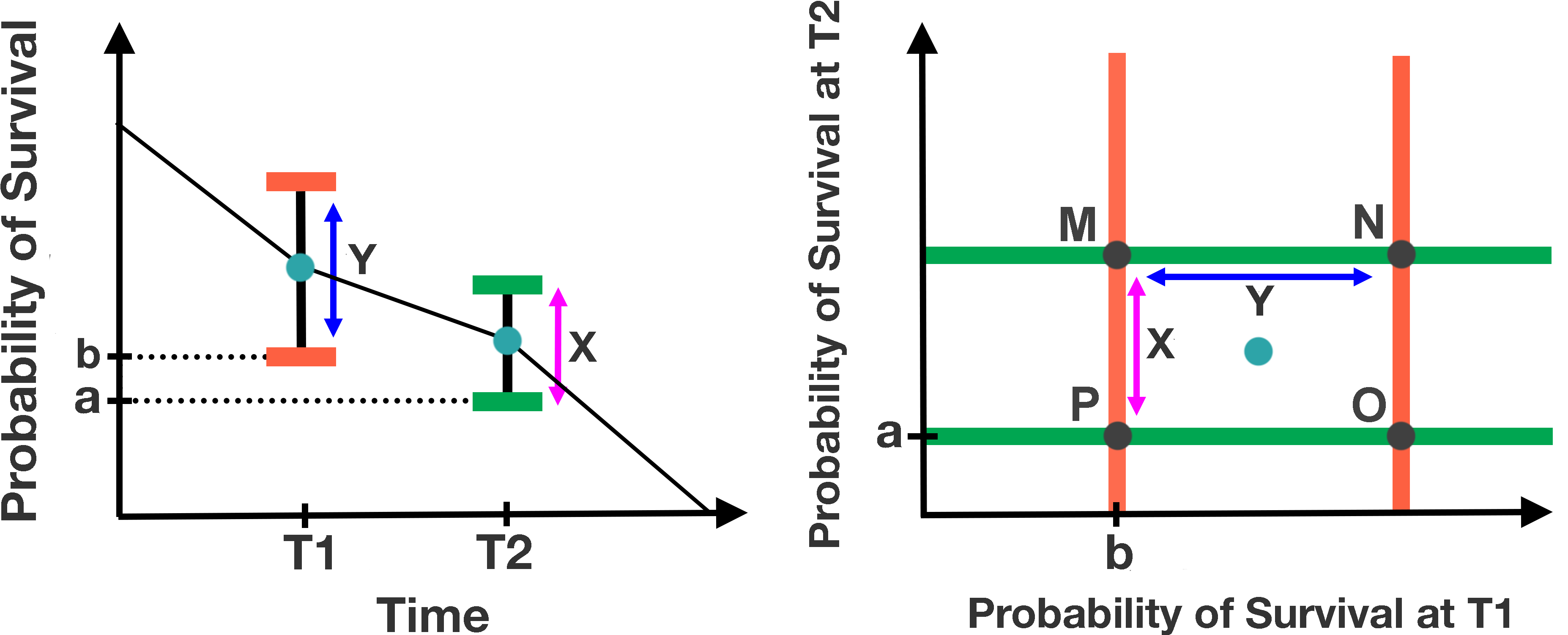}
  \caption{(left) An example of a two-discretized 
  survival graph.
  (right) The same survival curve viewed in $[0, 1]^2$.
  On a discretized survival graph, a survival curve is given as a sequence
  of points (shown in teal).
  In $[0, 1]^2$, these points are considered as a single point (shown
  in teal).
  Similarly, the error bars on the discretized survival graph correspond
  to the hyperplanes in $[0,1]^2$ of the same color.
  The simultaneous intervals on the survival graph 
  correspond to the orthotope \textbf{MNOP}.}
  \label{fig:vis-space}
\end{figure}

When dealing with $n$-discretized survival space, we can
view simultaneous prediction intervals as an
orthotope\footnote{An orthotope is a Cartesian product of intervals.} in 
$[0, 1]^n$.
We display the visual relationship between simultaneous prediction
intervals as viewed on a survival graph and as viewed as a subset of
$[0, 1]^n$ for a two-discretized survival curve in Figure~\ref{fig:vis-space}.
SPIEs
can be viewed as estimating an n-dimensional orthotope 
that, with the prescribed probability, 
contains a point sampled from the appropriate n-dimensional distribution.
From this perspective, given samples $\{\mathbf{z}^{(i)}\}$, Olshen's method yields orthotopes 
$O_k(\{\mathbf{z}^{(i)}\})$ of the form
\[
\prod_t [\mu(\{\mathbf{z}^{(i)}_t\}) - k \sigma(\{\mathbf{z}^{(i)}_t\}), 
\mu(\{\mathbf{z}^{(i)}_t\}) 
+ k  \sigma(\{\mathbf{z}^{(i)}_t\})],
\]
where $\mu$ gives the sample mean, $\sigma$ gives the sample standard deviation, and subscript $t$
gives the $t$th component.
In our context $\{\mathbf{z}^{(i)}\} = \{s(\mathbf{x}, \mathbf{m}^{(i)})\}$ 
is a set of survival curves specific to a patient
with features $\mathbf{x}$ (as shown by the green, red and blue curves in the
estimation phase of Figure \ref{fig:inference})
and $\{\mathbf{z}^{(i)}_t\} = \{s(\mathbf{x}, \mathbf{m}^{(i)})_t\}$ 
is the values of these survival curves at time $t$.
Ideally $k$ is chosen such that the orthotope 
contains a sample from the underlying
distribution with probability $1 - \alpha$.
To approximate this choice, Olshen's method bootstraps the samples to construct a 
collection of sample sets $\{\{\mathbf{z}^{(i)}\}_b\}_{b \in B}$.
Letting $\rho(C, D)$ denote the proportion of elements of $C$
that are elements of $D$,
Olshen's method chooses the smallest $k$ such that
\[
\mu\left(\left\{\rho\left(\{\mathbf{z}^{(i)}\}_b, 
O_k(\{\mathbf{z}^{(i)}\}_b)\right)\right\}_{b \in B}\right) \geq 1 - \alpha.
\]

\begin{figure}
\centering
  \includegraphics[width=\columnwidth]{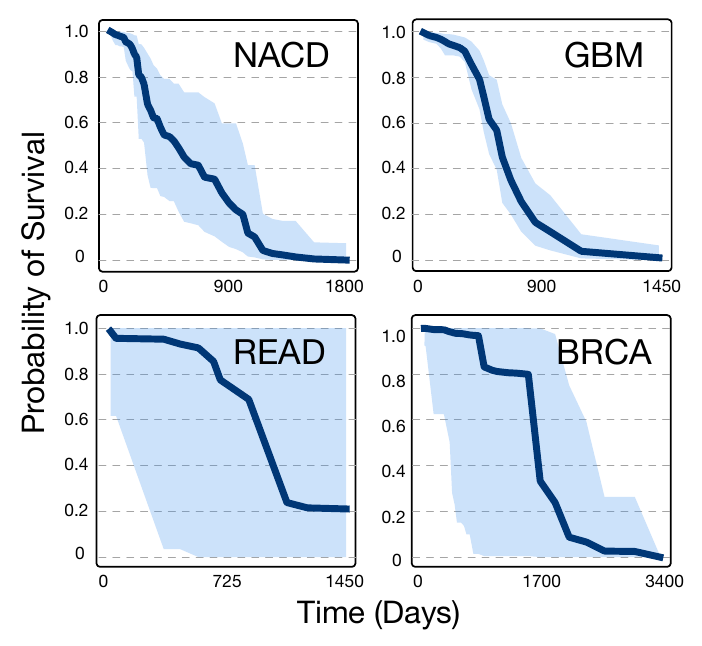}
  \caption{
  Examples of simultaneous 95\% prediction intervals 
  (estimated by GSPIE)
  for survival curve predictions from MTLR
  for patients from four different datasets.
  Due to the high censorship rates
  in READ and BRCA, the prediction intervals are very large, indicating
  that the predicted survival curve may be unreliable.
   }
  \label{fig:eg}
\end{figure}
\section{Methodology} 
\label{sec:Methodology}

The pipeline examined in this paper has two phases. In the
sampling phase (see Figure \ref{fig:inference}; left) 
bootstrap or posterior samples of the model
are acquired. These model samples approximate the uncertainty
that is present in the system and are used to quantify
uncertainty of predictions for new patients.
In the estimation phase (see Figure \ref{fig:inference}; right), 
given a new patient,
these model samples are used for
simultaneous prediction interval estimation.
In addition to investigating this framework in the context of
patient-specific survival analysis, 
this paper also introduces two alternative
SPIEs to that used by Olshen. 
One is a small modification to Olshen's method that
can produce asymmetric intervals; the other
takes a simple but effective greedy hill
climbing approach.

\subsection{Two-Sided Olshen's Method}

One disadvantage of Olshen's original method is that
the prediction intervals are constrained to be symmetric
about the mean. In the case of a highly asymmetric
sample distribution, this could be a costly restriction.
To address this, two-sided Olshen's method replaces
$O_k(\{\mathbf{z}^{(i)}\})$ by $O^{\pm}_k(\{\mathbf{z}^{(i)}\}$,
for both orthotope construction and the computation of $k$, where 
$O^{\pm}_k(\{\mathbf{z}^{(i)}\}$ is given by
\[
\prod_t [m(\{\mathbf{z}^{(i)}_t\}) - k \sigma^-(\{\mathbf{z}^{(i)}_t\}), 
m(\{\mathbf{z}^{(i)}_t\}) 
+ k  \sigma^+(\{\mathbf{z}^{(i)}_t\})],
\]
where $m$ gives the median, $\sigma^+$ gives the sample
root mean squared difference
between the median and the values greater than or equal to the median,
and $\sigma^-$ gives the sample 
root mean squared difference between the median and the
values less than or equal to the median. We think of $\sigma^+$
and $\sigma^-$ as capturing information about the variance of each side of
the distribution.

\subsection{Greedy Hill Climbing}

A simple alternative to Olshen's method is to do greedy 
optimization over the landscape of orthotopes.
We call this approach greedy simultaneous prediction intervals estimator (GSPIE).
GSPIE begins with an orthotope $O$ containing all
samples $\{\mathbf{z}^{(i)}\}$.
At each time step, GSPIE retracts one ``wall" of the orthotope
$O$ from its current
position inwards such that it lies on the nearest sample in 
$\{\mathbf{z}^{(i)}\}$ in the interior of $O$ (see
Figure \ref{fig:inference}; (7)).
GSPIE makes this decision greedily -- at each time step
it considers retracting each wall, 
selecting the retraction corresponding to the greatest reduction
in interval width per sample excluded.
For example, if exactly one sample lies on each wall,
GSPIE takes a step that reduces the sum over interval 
widths  $\sum_t O_{u_t}  - O_{\ell_t}$
of $O = \prod_t [O_{\ell_t}, O_{u_t}]$ by the value
\[\max\left(\max_t \{r_{\ell_t}(O)_{\ell_t} - O_{\ell_t}\}, 
\enspace \max_t \{O_{u_t} - r_{u_t}(O)_{u_t} \}\right),\]
where $r_x$ is the retraction operator discussed above 
moving ``wall" $x$ inwards.
The hill climbing procedure ends when the next retraction 
would leave $O$ containing less
than $1 - \alpha$ of a validation set 
$\{\mathbf{z}_{\mbox{\tiny val}}^{(i)}\}$.

\begin{figure*}
  \includegraphics[width=\linewidth]{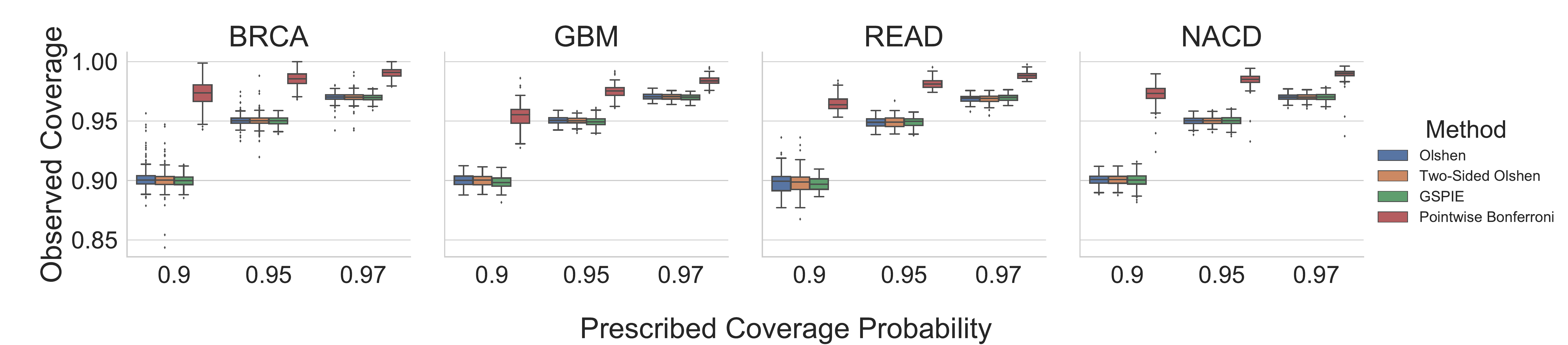}
  \caption{An accurate method's
  observed coverage should closely correspond to the prescribed coverage. Both of
  Olshen variants and GSPIE show much better accuracy than a naive Bonferroni
  correction, which guarantees conservative simultaneous intervals.}
\label{fig:val}
\end{figure*}

\section{Experiments} 
\label{sec:Exps}

In our experiments, we consider four survival datasets. 
The Northern Alberta
Cancer Dataset (NACD; 2402 patients, 53 features, 36\% censorship), 
includes patients with many types of cancer:
including lung, colorectal, head and neck, esophagus, stomach, etc.
The other three datasets are from 
The Cancer
Genome Atlas (TCGA) Research Network~\cite{tcga}:  
Glioblastoma multiforme
(GBM; 592 patients, 12 features, 18\% censorship),
Rectum adenocarcinoma (READ; 170 patients, 18 features, 84\%
censorship), 
and Breast invasive carcinoma (BRCA; 1095 patients, 61 features, 86\%
censorship).
These datasets contain ``right-censored patients'' --
patients for whom the dataset specifies only a lower-bound on 
survival time. 

For each dataset, we converted categorical variables into one-hot
labels and imputed the mean for missing real-valued features.
Each feature was normalized to mean zero and unit variance.
Each dataset was shuffled and divided into a training set and 
a testing set with a 75/25 split.
For its greedy hill climbing procedure, GSPIE divided each test
patient's survival
curve samples into an optimization set and a validation set with a 50/50 split.
Since our experiments are in the context of survival analysis,
we project the upper and lower bounds of each estimated SPI
into $\mathbb{S}^n$.

In Figure \ref{fig:eg} we show an example of 95\% simultaneous
prediction intervals (with linear interpolation) of a 
patient's survival function for each dataset. 
We attempted to select representative examples for each dataset. 
As might be expected, models trained on datasets with high 
censorship rates (READ and BRCA) yield highly uncertain survival functions.

\subsection{Evaluating Performance}

In evaluating SPIEs, there are two metrics of 
concern. First, an estimator should be accurate -- prescribing a 
$1 - \alpha$ 
coverage probability should yield simultaneous intervals 
that include a random sample with probability $1 - \alpha$.
Second, an estimator should produce tight prediction intervals -- \ie
simultaneous prediction intervals with small average width.
However, note that these two metrics are not independently meaningful.
It is neither useful to have loose, accurate intervals, nor to
have tight inaccurate intervals.\footnote{Though 
in many contexts we may not mind a method 
producing conservative simultaneous intervals, insofar as
conservatism does not come at the cost of tightness.}
Unfortunately, as far we are aware, there exists no established metric
that unifies accuracy and tightness in a manner that is robust to different
tradeoff preferences. Therefore, while we present these metrics separately,
we encourage readers to take a holistic perspective.

To examine the performance of these methods, 
we used multi-task logistic regression (MTLR), 
which had the strongest performance of the methods examined
in \cite{survey}, as our ISD model.
We trained MTLR using a regularization factor of 1/2, which was tuned with
five-fold cross-validation and \(\sqrt{N}\) time points (where $N$ is the 
size of the training set), such that an equal number of events occurs
between each two time points,
as is recommended by \cite{mtlr}.
We considered a Bayesian approach --
we collected 10,000 posterior samples from the posterior $p(\mathbf{m} \mid D)$
using NUTS \cite{nuts} and 
estimated joint prediction intervals for the predictive
posterior distribution $p(s(\mathbf{x}, \mathbf{m}) \mid D)$ of test set
patients.
We used an isotropic Gaussian prior for MTLR's model parameters
with a variance corresponding with the tuned $\ell_2$ regularization factor.
While not strictly necessary, enforcing this correspondence causes the 
predictions from the fitted MTLR model 
to coincide with the
maximum a posteriori (MAP) estimate.

\begin{figure*}
  \includegraphics[width=\linewidth]{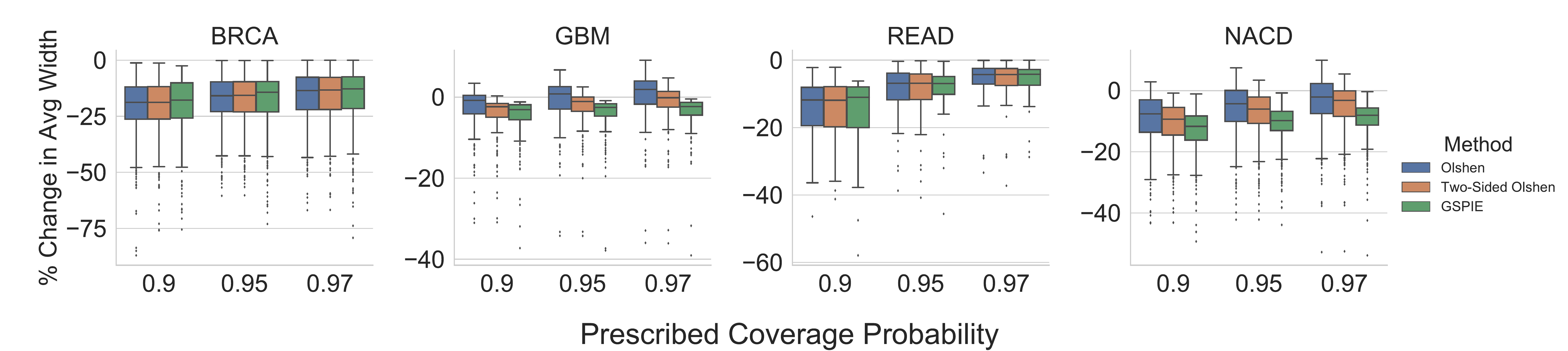}
  \caption{The figure shows the percent change in average width with respect
  to pointwise intervals with a Bonferroni correction -- lower is better. 
  All methods performed comparably on BRCA and READ 
  -- all were better than the baseline. However, on
  GBM and NACD, GSPIE 
  found tighter intervals than two-sided Olshen's method, 
  which itself found tighter intervals than Olshen's method.}
\label{fig:val-diff}
\end{figure*}

\subsubsection{Accuracy} 
\label{sec:valid}

To evaluate the accuracy of the prediction orthotopes,
we ran a second chain and collected 10,000 test samples from
the predictive posterior distribution $p(s(\mathbf{x}, \mathbf{m}) \mid D)$
for each test set patient.
Accurate $1-\alpha$ simultaneous prediction intervals should contain
roughly $(1 - \alpha) \times 10,000$ of these samples.
Figure~\ref{fig:val} displays the results of the analysis on 
each patient in the test set.
We include Olshen's method, two-sided Olshen's method, GSPIE, and
also a pointwise interval baseline\footnote{Pointwise intervals 
were obtained from percentile estimates via the ogive (\ie the empirical distribution 
with linear interpolation).} with a Bonferroni correction.
The most significant takeaway from the results is that 
the former three methods are much more accurate
than the conservative Bonferroni correction,
which guarantees a lower bound on the 
coverage probability of the simultaneous intervals.
Of course, the accuracy of all of the methods is highly 
dependent on the number of model samples that are
available.

\subsubsection{Tightness}
To evaluate the tightness of the prediction intervals
we measure the average interval width.
On a survival graph, this metric is proportional to
Lebesgue measure in the case that intervals exist 
for every time point, 
but extends more sensibly to our circumstance,
in which intervals exist for only a finite number of
time points.
Figure \ref{fig:val-diff} displays the result of the
analysis.
For each patient in the test set and for each
SPIE, 
we measured the percentage change in the average
interval width compared to pointwise intervals with a Bonferroni
correction -- lower is better.
For BRCA and READ,
all three methods appear to have
perform similarly, notably exceeding the baseline.
For GBM and NACD, there seems to be a clearer
distinction between the methods -- GSPIE 
generally gave tighter intervals than 
two-sided Olshen's method, which itself generally
gave tighter intervals than Olshen's method.

\begin{figure}
\centering
  \includegraphics[width=\columnwidth]{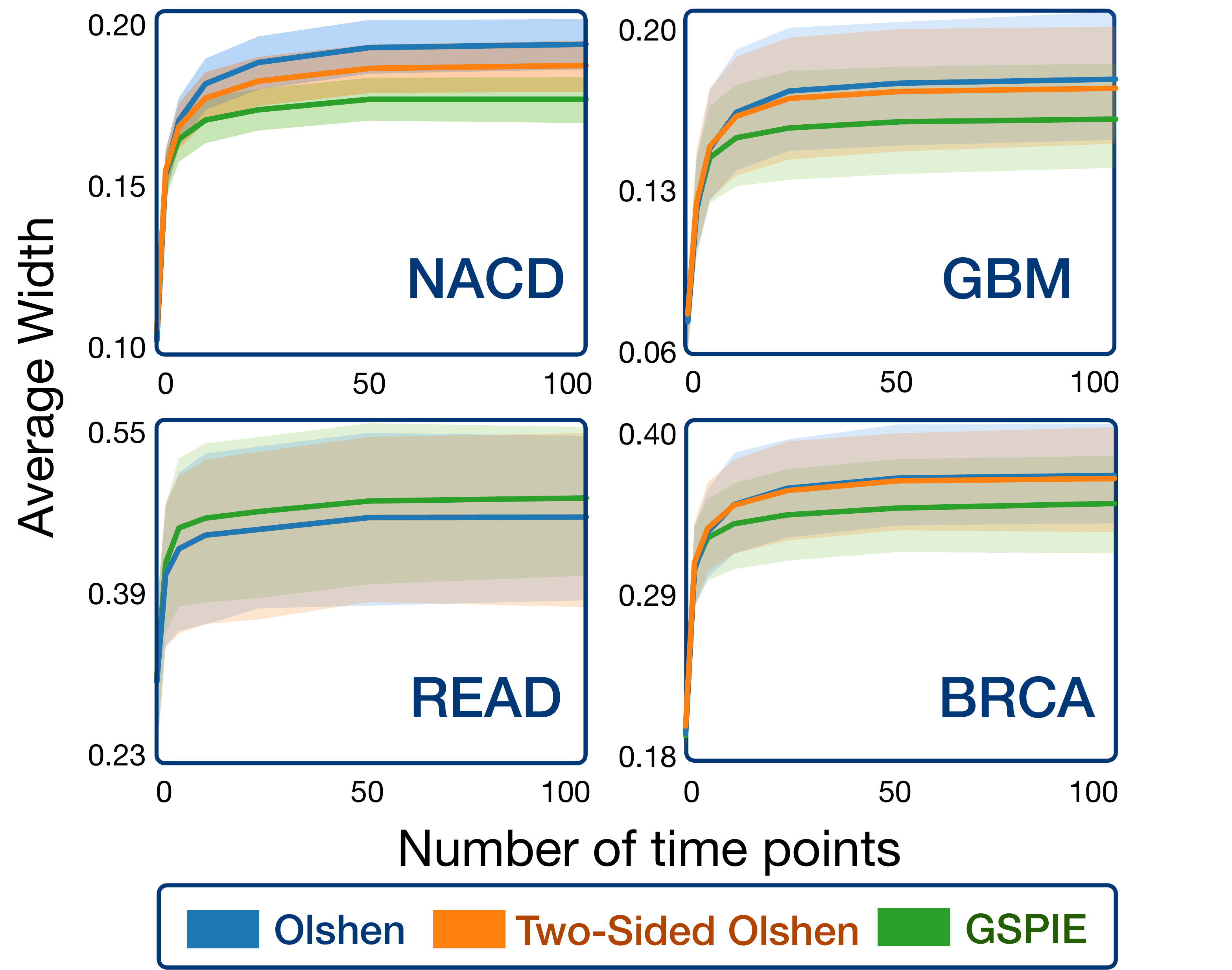}
  \caption{
  The figure shows SPI tightness as a function of discretization. 
  Even with fine discretizations, all three methods found highly
  nontrivial prediction intervals.}
  \label{fig:num_bins}
\end{figure}

\subsection{Discretization of Continuous Time Curves}

Although the methods discussed in this paper
are limited to a finite number of time points,
there is no reason they cannot be applied
to ISD models that produce continuous time
curves. However, one might worry that 
using a very large number of time points (\ie a fine discretization) 
might cause SPIEs to produce trivially large prediction intervals. 
To investigate this possibility, we used the Weibull
accelerated failure time model \cite{cox-cp-aft}, 
which produces continuous time patient-specific
survival curves, on each of the four datasets.
We considered a Bayesian perspective, assuming
an isotropic Gaussian prior  
for the model parameters and a Gumbel distribution
with a half-normal hyperprior for the noise. 
We show the results in Figure \ref{fig:num_bins}, where
mean average width with 95\% pointwise confidence intervals
is plotted as a function of the discretization for each
of the different datasets.
We observe that all three methods were able to find
highly nontrivial simultaneous prediction intervals
for finely discretized curves. 
We expect this result to hold generally
across other continuous models and datasets.

\subsection{Discussion} 
\label{sec:Discussion}

Our experiments suggest that Olshen's method, two-sided Olshen's
method, and GSPIE all perform well as SPIEs. 
All three are accurate, efficient, 
find relatively tight intervals, and can be applied
to continuous time models with fine discretizations.
The right choice of estimator appears to be context dependent.
For example, with MTLR, on NACD and GBM, GSPIE may offer the best
compromise between accuracy and tightness,
whereas on READ and BRCA, it is less clear. Luckily, given
that all three methods are efficient and can make use of
the same model samples, it is relatively easy to try and
test\footnote{One can simply take more model samples to
verify that estimated intervals meet their prescription,
as was done in our experiments.} all of them.

\section{Conclusion}

In this work we promote a framework for quantifying the uncertainty
of patient-specific survival curves. 
The methods examined here are easily-implementable 
and applicable to a large class of ISD models.
We hope this framework will allow clinicians to more comfortably
consider ISD models as a source of information for
patient-specific decisions.

Lastly, although the focus of this paper is patient-specific
survival analysis, the SPIEs we examined
are general purpose tools for estimating
SPI from samples. We anticipate that these methods could
prove useful in many contexts outside of survival analysis,
such as time series analysis, trigonometric regression, 
and quantile regression.

\bibliography{ijcai19}

\end{document}